\documentclass[sigconf]{acmart}
\makeatother
\usepackage{booktabs} 
\setcopyright{rightsretained}

\acmDOI{}

\acmISBN{}

\acmPrice{}

\acmConference[MAICS'17]{The 28th Modern Artificial Intelligence and Cognitive Science Conference}{March 2017}{Purdue University Fort Wayne} 
\copyrightyear{2017}

\begin{document}
\title{Learning Photography Aesthetics with Deep CNNs}
\author{Gautam Malu}
\email{gautam.malu@research.iiit.ac.in}
\authornote{Corresponding author}
\affiliation{%
  \institution{International Institute of Information Technology}
  \city{Hyderabad} \state{India} 
}
\author{Raju S. Bapi}
\email{raju.bapi@iiit.ac.in}
\affiliation{%
\institution{International Institute of Information Technology  \\
\& University of Hyderabad}
  \city{Hyderabad} \state{India} 
}
\author{Bipin Indurkhya}
\email{bipin@iiit.ac.in}
\affiliation{%
\institution{International Institute of Information Technology}
  \city{Hyderabad} \state{India} 
 }

\renewcommand{\shortauthors}{Gautam et al.}

\begin{abstract}
Automatic photo aesthetic assessment is a challenging artificial intelligence task. Existing computational approaches have focused on modeling a single aesthetic score or class (good or bad photo), however these do not provide any details on why the photograph is good or bad; or which attributes contribute to the quality of the photograph.
To obtain both accuracy and human-interpretability, we advocate learning the aesthetic attributes along with the prediction of the overall score. For this purpose, We propose a novel multi-task deep convolution neural network (DCNN), which jointly learns eight aesthetic attributes along with the overall aesthetic score. We report near-human performance in the prediction of the overall aesthetic score. To understand the internal representation of these attributes in the learned model, we also develop the visualization technique using back propagation of gradients. These visualizations highlight the important image regions for the corresponding attributes, thus providing insights about model's understanding of these attributes. We showcase the diversity and complexity associated with different attributes through a qualitative analysis of the activation maps.
\end{abstract}
\maketitle
\section{Introduction}
Aesthetics is the study of science behind the concept and perception of beauty. Although aesthetics of photograph is subjective, some aspect of its depends on the standard photography practices and general visual design rules. With the ever increasing volume of digital photographs, automatic aesthetic assessment is becoming increasingly useful for various applications, such as a personal photo assistant, photo manager, photo enhancement, image retrieval etc. Conventionally, automatic aesthetic assessment tasks have been modeled as either a regression problem (single aesthetic score) ~\cite{kao2015visual, wu2011learning, kao2016hierarchical} or as a classification problem (aesthetically good or bad photograph) ~\cite{datta2006studying,lu2014rapid,lu2015deep}.      

\begin{figure*}
  \includegraphics[width=\textwidth]{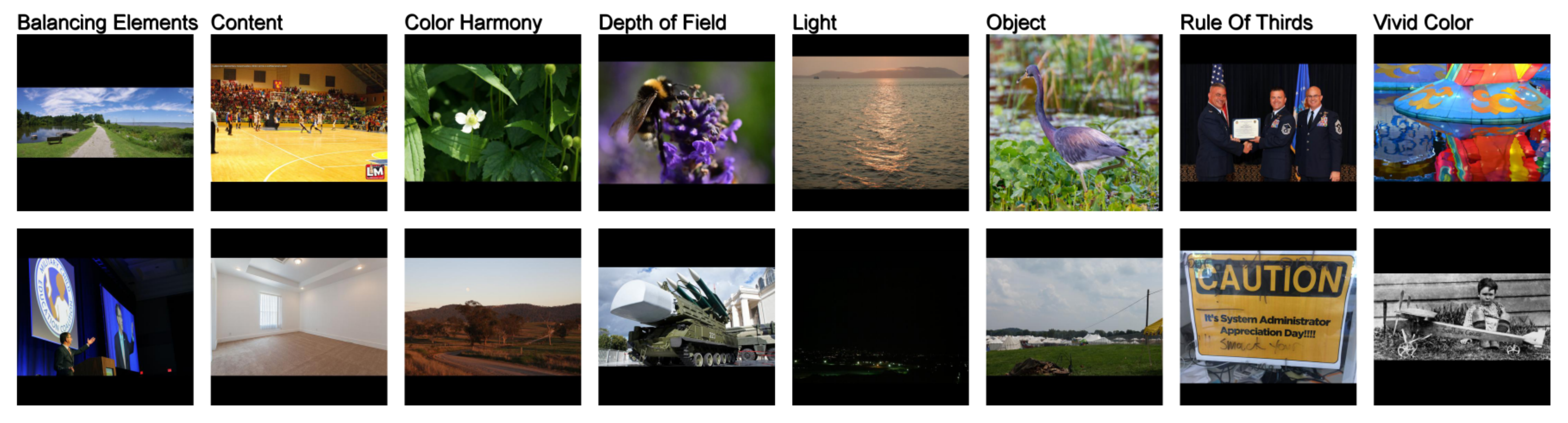}
  \caption{Sample images taken from AADB for each attribute. top row: Highest rated images, bottom row: Lowest Rated Images. All images were padded to maintain aspect ratio for illustration purposes.}
  \label{fig:sample}
\end{figure*}

Intensive data driven approaches have made substantial progress in this task, although it is a very subjective and context dependent task. Earlier approaches used custom designed features based on photography rules (e.g., focus, color harmony, contrast, lighting, rule of thirds) and semantic information (e.g., human profile, scene category) from low level image descriptors (\textit{e.g.} color histograms, wavelet analysis) \cite{datta2006studying,ke2006design,dhar2011high,sun2009photo,luo2008photo,luo2011content,nishiyama2011aesthetic,lo2012statistic,bhattacharya2010framework} and generic image descriptors ~\cite{marchesotti2011assessing}. With the  evolution of deep learning based techniques, recent approaches have introduced deep convolution neural networks (DCNN) in aesthetic assessment tasks ~\cite{lu2014rapid,lu2015deep,tian2015query,kao2015visual,kong2016photo}.  

Although these approaches give near-human performance in classifying whether a photograph is ``good" or ``bad", they do not give detailed insights or explanation for such claims. For example, if a photograph received a bad rating, one would not get any insights about the attributes (e.g., poor lighting, dull colors etc.) that led to that rating. We propose an approach in which we identify (eight) such attributes (such as Color Harmony, Depth of Field etc.) and report those along with the overall score. For this purpose, we propose a multi-task deep convolution network (\emph{DCNN}) which simultaneously learns the eight aesthetic attributes along with the overall aesthetic score. We train and test our model on the recently released aesthetics and attribute database (\emph{AADB}) ~\cite{kong2016photo}. Following are the eight attributes as mentioned in \cite{kong2016photo} (Figure ~\ref{fig:sample}):
\begin{enumerate}
\item \emph{Balancing Element} - Whether the image contains balanced elements.
\item \emph{Content} - Whether the image has good/interesting content.
\item \emph{Color Harmony} - Whether the overall color composition is harmonious. 
\item \emph{Depth of Field} - Whether the image has shallow depth of field.
\item \emph{Light} - Whether the image has good/interesting lighting.
\item \emph{Object Emphasis} - Whether the image emphasizes foreground objects.
\item \emph{Rule of Thirds} - Whether the image follows rule of thirds principle. The rule of thirds involves dividing the photo into 9 parts with 2 vertical and 2 horizontal lines. The important elements and leading lines are placed on/near these these lines and intersections of these lines.
\item \emph{Vivid Color} - Whether the image has vivid colors, not necessarily harmonious colors.
\end{enumerate}

We also develop attribute activation maps (Figure ~\ref{fig:visualization}) for visualization of these attributes. These maps highlight the salient regions for the corresponding attribute, thus providing us insights about the representation of these attributes in our trained model.
\newline
In summary, followings are the main contributions of our paper:
\begin{enumerate}
\item We propose a novel deep learning based approach which simultaneously learns eight aesthetic attributes along with the overall score. These attributes enable us to provide more detailed feedback on automatic aesthetic assessment. 
\item We also develop localized representation of these attributes from our learned model. We call these \emph{attribute activation maps} (Figure ~\ref{fig:visualization}). These maps provide us more insights about model's interpretability of the attributes. 
\end{enumerate}

\section{Related Work}
Most of the earlier works have used low-level image features to design high level aesthetic attributes as mid-level features and trained aesthetic classifier over these features. Datta \textit{et al.} \cite{datta2006studying} proposed 56 visual features based on standard photography and visual design rules to encapsulate aesthetic attributes from low-level image features. Dhar \textit{et al.} divided aesthetic attributes into three categories Compositional (\textit{e.g.} Depth of field, Rule of thirds), Content(\textit{e.g.} faces, animals, scene types), Sky-Illumination (\textit{e.g.} clear sky, sunset sky). They trained individual classifiers for these attributes from low-level features (\textit{e.g.} color histograms, center surrounding wavelets, haar features) and used outputs of these classifiers as input features for the aesthetic classifier. 

Marchesotti \textit{et al.} \cite{marchesotti2015discovering}, proposed to learn aesthetic attributes from textual comments on the photographs using generic image features. Despite increased performance, many of these textual attributes (good, looks great, nice try) do not map to well-defined visual characteristics.  Lu \textit{et al.} ~\cite{lu2014rapid} proposed to learn several meaningful style attributes, and used these to fine-tune the training of aesthetics classification network. Kong \textit{et al.} \cite{kong2016photo} proposed attribute and content adaptive DCNN for aesthetic score prediction.

\begin{figure}
  \includegraphics[width=\linewidth]{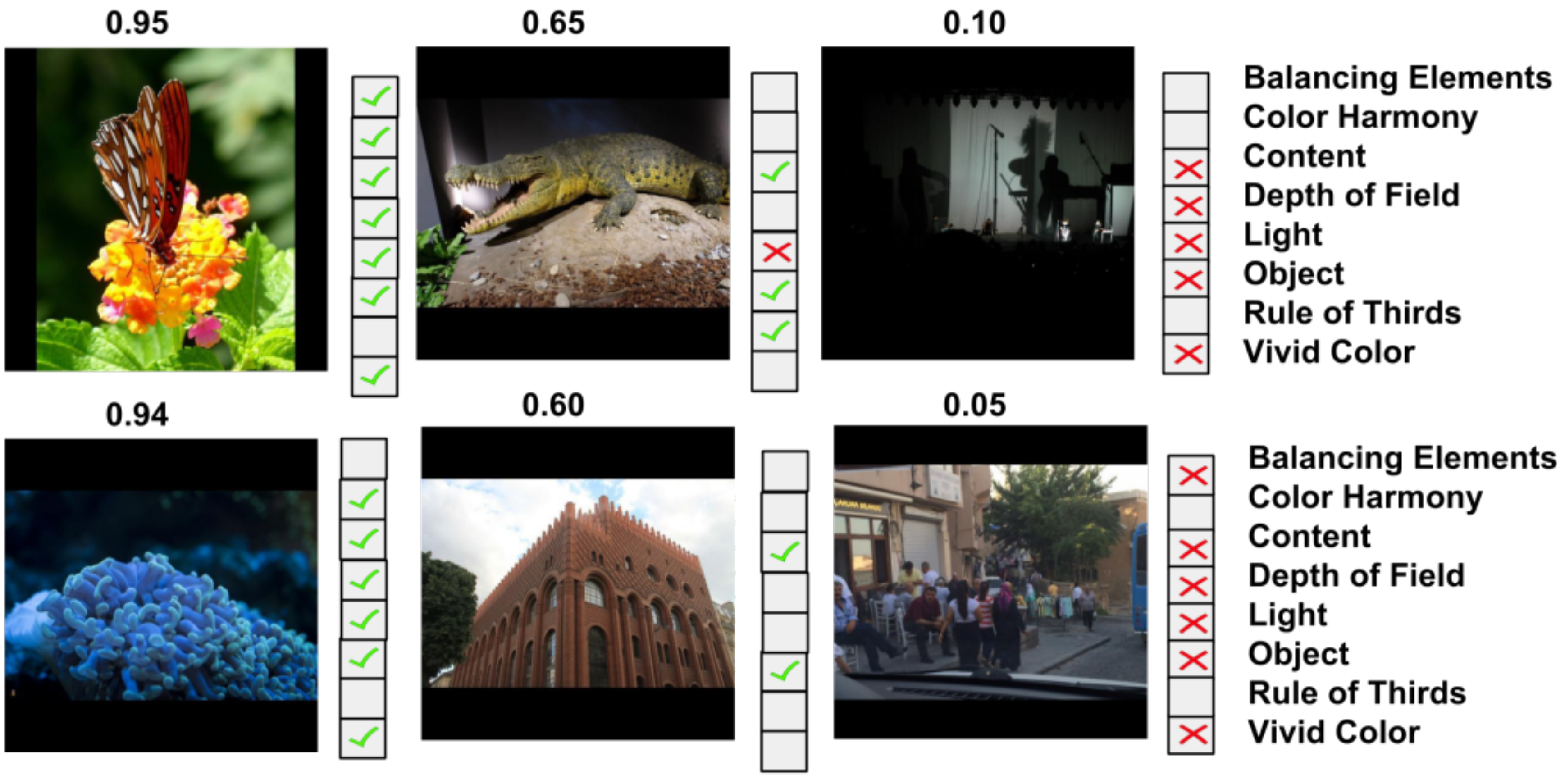}
  \caption{Sample images from AADB testing data. first column: Images rated high on aesthetic score, second Column: Images rated at mid-level, third Column: Images rated low.}
  \label{fig:good_bad_ugly}
\end{figure}

However, none of the previous works report the aesthetic attributes themselves. These attributes are used as features to predict the overall aesthetic score/class. In this paper, we learn aesthetic attributes along with the overall score, not just as intermediate features but as auxiliary information. Aesthetic assessment is relatively easier in images with evident high and low aesthetics than in ordinary images with marginal aesthetics (Figure ~\ref{fig:good_bad_ugly}). For these images, attributes information would greatly supplement the quality of feedback from an automatic aesthetic assessment system. 

Recently, deep learning techniques have shown significant performance gains in various computer vision tasks such as object classification, localization ~\cite{krizhevsky2012imagenet, Simonyan14c,szegedy2015going}. In deep learning, non-linear features are learned in a hierarchical fashion in increasing complexity 
\small(e.g. colors, edges, objects\small). The aesthetic attributes can be learned as combinations of these features. Deep learning techniques have shown significant performance gains in comparison with traditional machine learning approaches for aesthetic assessment tasks ~\cite{lu2014rapid, lu2015deep, tian2015query, kao2015visual, kong2016photo}. Unlike traditional machine learning techniques, features are also learned during training in deep learning techniques. However these internal representations of DCNNs are still opaque. Various visualization techniques ~\cite{zeiler2014visualizing, zhou2014object, mahendran2015understanding, dosovitskiy2015inverting, zhou2016learning, selvaraju2016grad} have been proposed to visualize the internal representations of DCNNs in an attempt to have a better understanding of their working. However, these visualization techniques have not been applied in aesthetic assessment tasks. In this article, we apply the gradient based visualization technique proposed by Zhou \textit{et al.} ~\cite{zhou2016learning} to obtain attribute activation maps. These maps provide localized representation of these attributes. Additionally we also apply similar visualization technique \cite{selvaraju2016grad} to the model provided by Kong \textit{et al.} \cite{kong2016photo} to obtain similar maps for qualitative comparison of our results with the earlier approach.

\section{Method}
\subsection{Architecture}
We use the deep residual network (ResNet50)~\cite{he2016deep} to train all the
attributes along with the overall aesthetic score. ResNet50 has 50 layers which can be divided into 16 successive residual blocks. Each residual block contains 3 convolution layers followed by the batch normalization layer (Figure ~\ref{fig:visualization}). Each residual block is followed by a rectified linear activation layer (ReLU) \cite{nair2010rectified}. We take these rectified convolution maps from the ReLU output of all these 16 residual blocks, and pool features from each of these 16 blocks with a global average pooling (GAP) layer. GAP layer gives the spatial average of these rectified convolution maps. Then we concatenate all these pooled features and use this as a feature for a fully connected layer which produces the desired outputs (aesthetic attributes and the overall score) as shown in Figure ~\ref{fig:visualization}. We model the attribute and score prediction  as a regression problem with mean squared error as loss function. 
Due to this simple connectivity structure, we are able to identify the importance of image regions by projecting the weights of the output layer on to the rectified convolution maps, a technique we call \emph{attribute  activation mapping}. This technique was first introduced by Zhou \textit{et al.}~\cite{zhou2016learning} to get class activation maps for different semantic classes in image classification task.  

\subsection{Attribute Activation Mapping}
\begin{figure}
  \includegraphics[width=\linewidth]{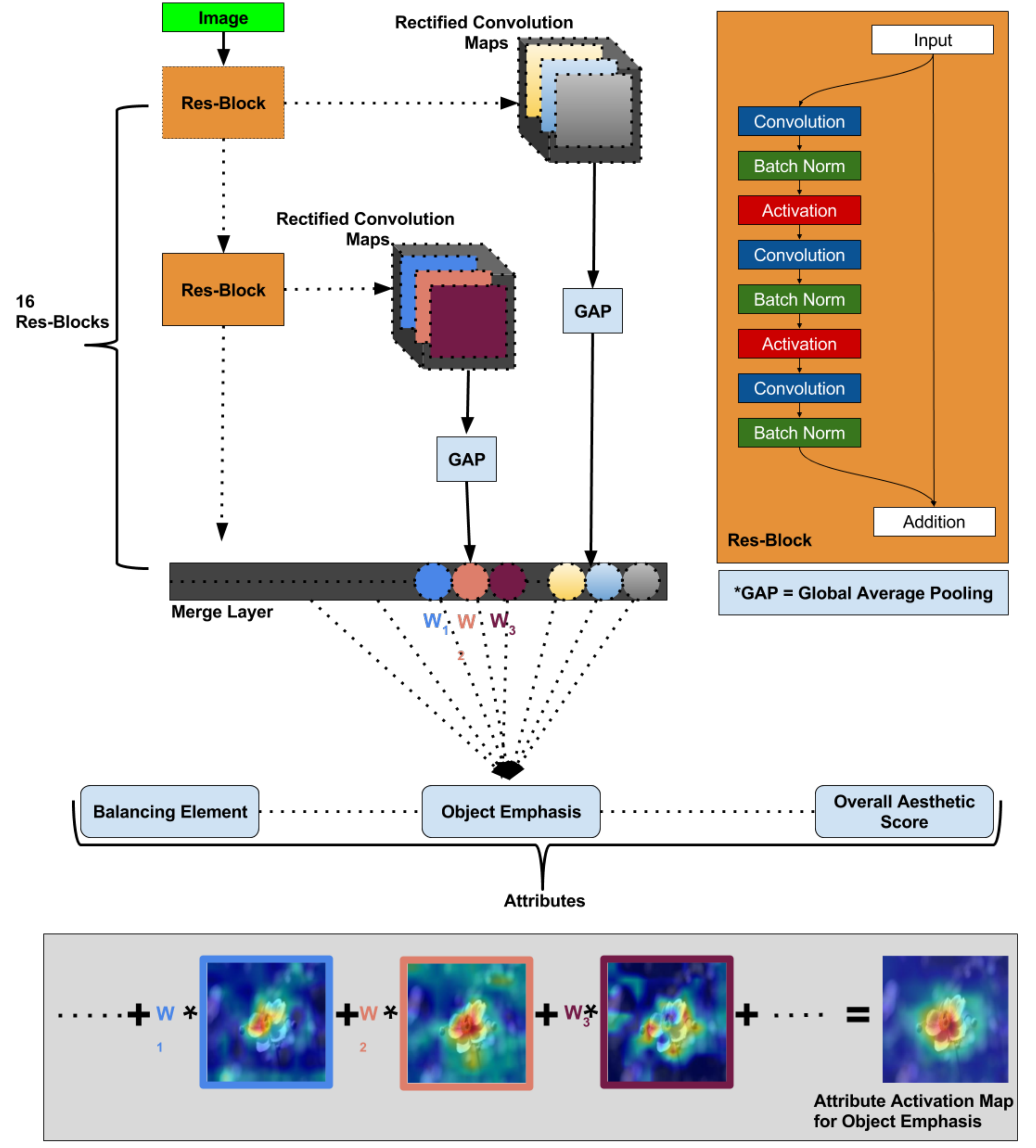}
  \caption{Our approach for generating attribute activation maps. The predicted score for a given attribute (object emphasis in the figure) is mapped back to the rectified convolution layers to generate the attribute activation maps. These maps highlight the attribute-specific discriminative regions as shown in the bottom section.}
  \label{fig:visualization}
\end{figure}

For a given image, let \textit{\emph{$f_k$\small($x, y$\small)}} represent the activation of unit \textit{$k$} in the rectified convolution map at spatial location \textit{\small(x, y\small)}. Then, for unit k, the result of performing global average pooling is $F$\textsuperscript{k} = $\sum\nolimits_{\textit{x,y}} \textit{\emph{$f_k$\small($x, y$\small)}}$. Thus, for a given attribute \textit{a}, the input to the regression layer, \textit{$R_a$}, is $\sum\nolimits_{\textit{x}} \textit{\emph{$w^a_k$\textit{$F_k$}}}$ where \textit{$w^a_k$} is the weight corresponding to attribute \textit{a} for unit k. Essentially, \textit{$w^a_k$} indicates the importance of \textit{$F_k$} for attribute \textit{a} as shown in Figure ~\ref{fig:visualization}.

We also synthesized similar attribute maps from the model proposed by Kong \textit{et al.} ~\cite{kong2016photo}. We did not have the final attribute and content adapted model from ~\cite{kong2016photo} due to patent rights but Kong \textit{et al.} shared the attribute adapted model with us. That model is based on alexnet architecture ~\cite{krizhevsky2012imagenet} consisting of fully connected layers along with convolution layers. In this architecture, outputs of convolution layers are separated from desired outputs by three stacked fully connected layers. The outputs from last FC layer are regression scores of attributes. In this architecture we compute weight of layer \emph{k} for attribute \emph{a} as summation of gradients (\emph{$g_k^a$}) of outputs with respect to $k^{th}$ convolution layer $w^a_k$ = $\sum\nolimits_{\textit{x,y}} \textit{\emph{$g_k^a$\small($x, y$\small)}}$.  
This technique was first introduced by Selvaraju \textit{et al.}~\cite{selvaraju2016grad} to get class activation maps for different semantic classes and visual explanation (answers for questions).    

\subsection{Implementation Details}
Out of $10000$ samples present in the AADB dataset, we have trained our model on $8500$ training samples. 500 and 1000 images have been set aside for validation and testing purposes, respectively. As the number of training samples \small($8500$\small) is not adequate for training of such a deep network \small(23,715,852 parameters\small) from scratch, we used a pre-trained ResNet50. It was trained on 1000-class Imagenet classification dataset ~\cite{deng2009imagenet} with approximately $1.2$ million images. We fixed the input image size to $299\times299$. We used \emph{horizontal flip} of the input images as a data augmentation technique. The last residual block gives convolution maps of size $10\times10$, so we reduce the sizes of the convolution maps from the previous Res-Blocks to the same size with appropriate sized average pooling. As ResNet50 has batch normalization layers, it is very sensitive to batch size. We fixed the batch size to 16 and trained it for 16 epochs. We report our model\textquotesingle s performance on test set ($1000$ images) provided in AADB. We have made our implementation publicly available~\footnote{\label{code_src}{\url{https://github.com/gautamMalu/Aesthetic_attributes_maps}}}.

\subsection{Dataset}
\begin{figure}[t]
\includegraphics[width=\linewidth]{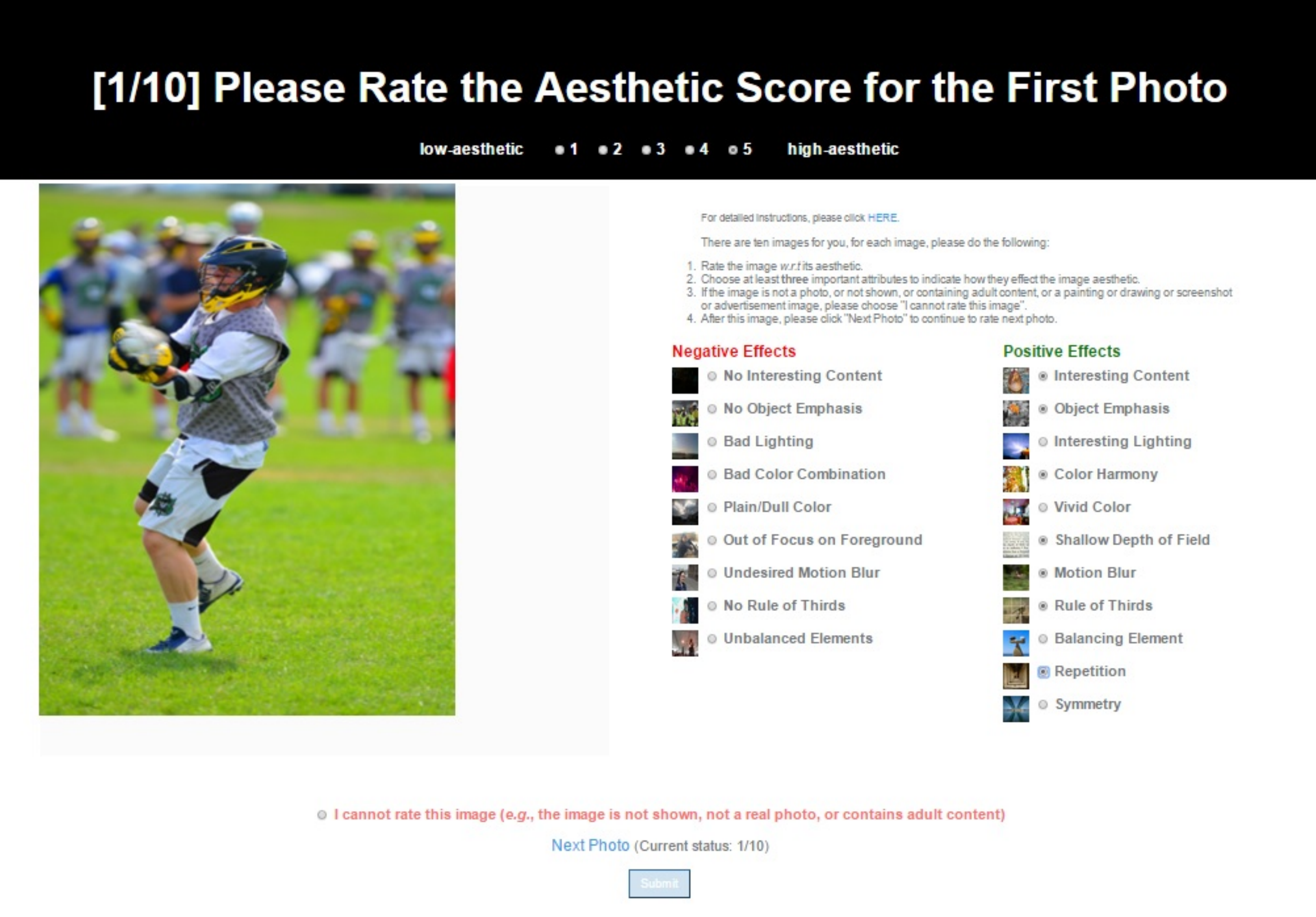}
\caption{Interface of data collection adopted by Kong \textit{et al.}\cite{kong2016photo}.}
\label{fig:interface}
\end{figure}

As mentioned earlier, we have used the aesthetics and attribute database (AADB) provided by Kong \textit{et al}.~\cite{kong2016photo}. AADB provides overall ratings for the photographs along with the ratings on the eleven aesthetic attributes as mentioned in ~\cite{kong2016photo} (Figure ~\ref{fig:sample}). 
Users were asked to provide information about the effectiveness of these attributes on the overall aesthetic score. For example, if \emph{object emphasis} is positively contributing towards the overall aesthetics of a photograph, user will give a score of \emph{+1} for the attribute, if object is not emphasized adequately and this is contributing negatively towards the overall aesthetic score of the photograph, user will give a score of \emph{-1} for the attribute (See  Fig~\ref{fig:interface}). The users also rated the overall aesthetic score on a scale of 1 to 5, with 5 being the most aesthetically pleasing score. Each image was rated by at least \emph{5} persons. The mean score was taken as the ground truth score for all attributes and the overall score.

\begin{figure}
  \includegraphics[width=\linewidth]{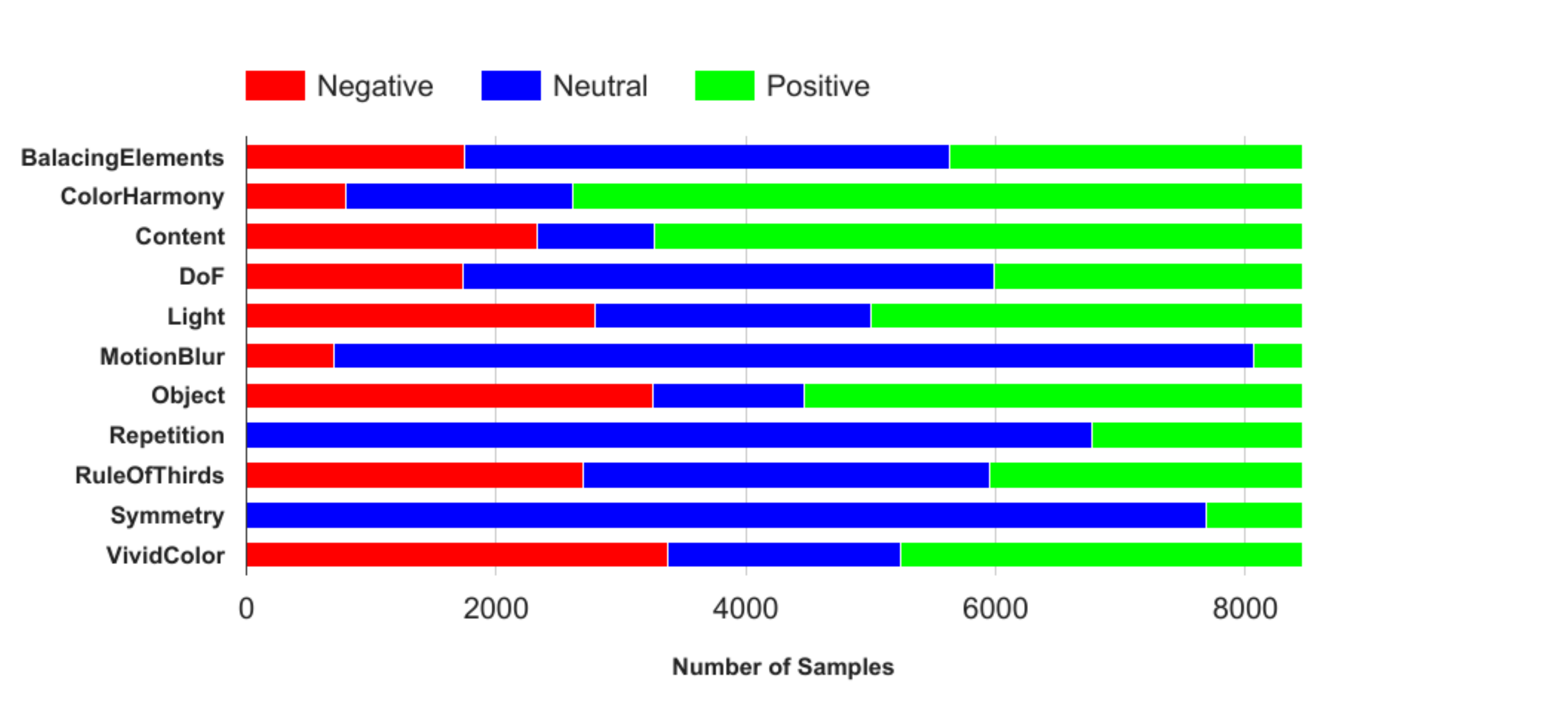}
  \caption{The distribution of all the eleven attributes in the training data of AADB. Most of the images are rated neutral for motion blur, repetition and symmetry.}
  \label{fig:attrDist}
\end{figure}

If an attribute has enhanced the image quality, it was rated positively and if the attribute has degraded the image aesthetics it was rated negatively. The default zero (null) means the attribute does not affect the image aesthetics. For example, positive \emph{vivid color} means the vividness of the color presented in an image has a positive effect on the image aesthetics; while the negative \emph{vivid color} means the image has dull color composition. All the attributes except for \emph{Repetition} and \emph{Symmetry} are normalized to the range of [-1, 1] \emph{Repetition} and \emph{Symmetry} are normalized to the range of [0, 1], as negative values are not justifiable for these two attributes.  The overall score is normalized to the range of [0, 1].Out of these eleven attributes, we omit \emph{Symmetry}, \emph{Repetition} and \emph{Motion blur} attributes from our experiment as most of the images rated null for these attributes (Figure ~\ref{fig:attrDist}).  We model the other eight attributes along with the overall aesthetic score as a regression problem.

\section{Results \& Discussion}
To evaluate the aesthetic attribute scores predicted by our model, we report the Spearman\textquotesingle s ranking correlation coefficient \small($\rho$\small) between the estimated aesthetic attribute score and the corresponding ground truth score for the testing data. The ranking correlation coefficient \small($\rho$\small) evaluates the monotonic relationship between estimated scores and ground truth scores, hence there is no need of explicit calibration between them. The correlation coefficient lies in the range of [-1, 1], with greater values corresponding to higher correlation and vice-versa. For baseline comparison, we also train a model by fine tuning a pre-trained ResNet50 and label it as ResNet50-FT. Fine-tuning here refers to modifying the last layer of the pre-trained ResNet50 \cite{he2016deep} and training it for our aesthetic attribute prediction task.
Table ~\ref{tab:results} lists the performance on AADB using the two approaches. We also report the performance of the model shared by Kong \textit{et al.}\cite{kong2016photo}.  
\begin {table}
\caption{Spearman\textquotesingle s rank correlations for all the attributes. All correlation coefficients ($\rho$) are significant at $p<0.0001$. The coefficients marked with a * are best results for respective attributes.} 
\label{tab:results}
\begin{center}
\begin{tabular} {|l|c|c|c|}
\hline
\emph{Attribute} & ResNet50-FT & Kong \textit{et al.}\cite{kong2016photo}& Our method\\
\hline
Balancing Elements & $0.184$ & \textbf{0.220*} & $0.186$ \\
\hline
Content & $0.572$ & $0.508$ & \textbf{0.584*} \\
\hline
Color Harmony & $0.452$ & $0.471$ & \textbf{0.475*} \\
\hline
Depth of Field & $0.450$ & $0.479$ & \textbf{0.495*} \\
\hline
Light & $0.379$ & \textbf{0.443*} & $0.399$ \\
\hline
Object Emphasis & $0.658$ & $0.602$ & \textbf{0.666*} \\
\hline
Rule of Thirds & $0.175$ & \textbf{0.225*} & $0.178$ \\
\hline
Vivid Colors & $0.661$ & $0.648$ & \textbf{0.681*} \\
\hline
Overall Aesthetic Score  & $0.665$ & $0.654$\footnotemark$/0.678$ & \textbf{0.689*} \\
\hline
\end{tabular}
\end{center}
\end{table}

\footnotetext{The $\rho$ reported by Kong \textit{et al.} ~\cite{kong2016photo} for their final content and attribute adaptive model is $0.678$, here we are reporting the performance of the model shared by them.}

It should be noted that the spearman\textquotesingle s coefficient between the estimated overall aesthetic score and the corresponding ground truth reported by Kong \textit{et al}.\cite{kong2016photo} was 0.678. They did not report any metrics for the other aesthetic attributes. They used ranking loss along with mean squared error as loss functions. Their final approach was also content adaptive. 
As can be seen from the results reported in Table~\ref{tab:results}, our model managed to outperform their approach in overall aesthetic score in-spite of only being trained with mean square error and without any content adaptive framework .  
Our model significantly underperformed for \emph{Rule of Thirds} and \emph{Balancing elements} attributes. These attributes are location sensitive attributes. \emph{Rule of thirds} deals with positioning of the salient elements, \emph{Balancing Elements} deals with relative positioning of objects with each other and the frame. In our model, due to use of global average pooling \small(GAP\small) layers after activation layers we are losing location specificity. We selected GAP layer to reduce the number of parameters. The number of training samples \small($8500$\small) allows learning of only small parameter space. We also warp the input images to the fixed size input \small($299x299$\small), thus destroying the aspect ratio. These could be possible reasons for the under-performance of the model for these compositional and location sensitive attributes. Across all the attributes, our proposed method reports better results than ResNet50 fine-tuned model. Our model performs better than the model provided by Kong {et al.}~\cite{kong2016photo} for five-out-of-eight attributes.

Aspects of aesthetic judgments are very subjective in nature. To quantify this subjectivity. In AADB the ground-truth score is the mean score of ratings given by different individuals. To quantify the agreement between ratings, $\rho$ between each individual\textquotesingle s ratings and the ground-truth scores was calculated. The average of $\rho$ is reported in Table ~\ref{tab:human_results}.  Our model actually outperforms the human consistently (as measured by $\rho$) averaged across all raters. However, when considering only the ``power raters'' who have annotated more images, human evaluators consistently outperform model\textquotesingle s results.

\begin{table}
\caption{Human performance on AADB. Our model actually outperforms the human consistently (as measured by $\rho$, last row) averaged across all raters (first row). However, when considering only the ``power raters'' who have annotated more images, human consistently outperform our model (second and third row).}
\label{tab:human_results}
\begin{center}
\begin{tabular} {|l|c|r|}
\hline
\emph{Number of Images rated} & Number of Raters & $\rho$ \\
\hline
$> 0$ & $195$& $0.6738$ \\
\hline
$> 100$ & $65$ & $0.7013$ \\
\hline
$> 200$ & $42$ & $0.7112$ \\
\hline
Our Approach & $-$ & $0.689$ \\
\hline
\end{tabular}
\end{center}
\end{table}

\begin{figure*}
  \includegraphics[width=\textwidth]{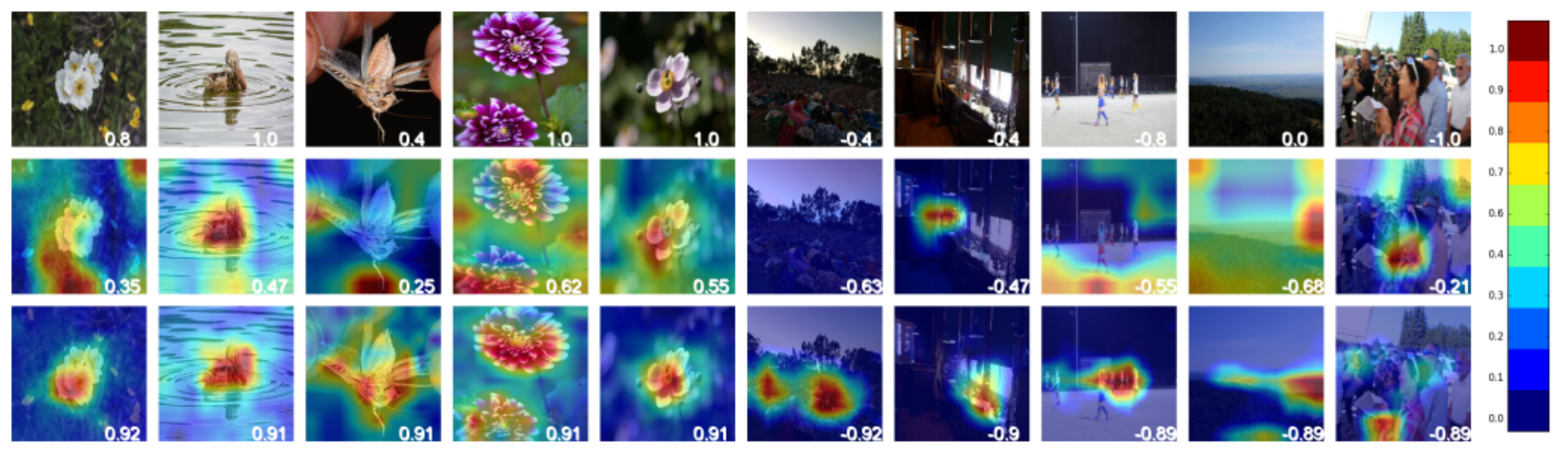}
  \caption{Object Emphasis activation maps. First row: Original Images \small(marked with ground truth score at the bottom right\small), second row: Activation Maps from Kong \textit{et al.}~\cite{kong2016photo} model~\small(marked with predicted score from their given model\small), third row: Activation Maps from our method~\small(marked with our predicted score \small). Color-bar indicates the color encoding of activation.}
  \label{fig:object}
\end{figure*}

\begin{figure*}
  \includegraphics[width=\textwidth]{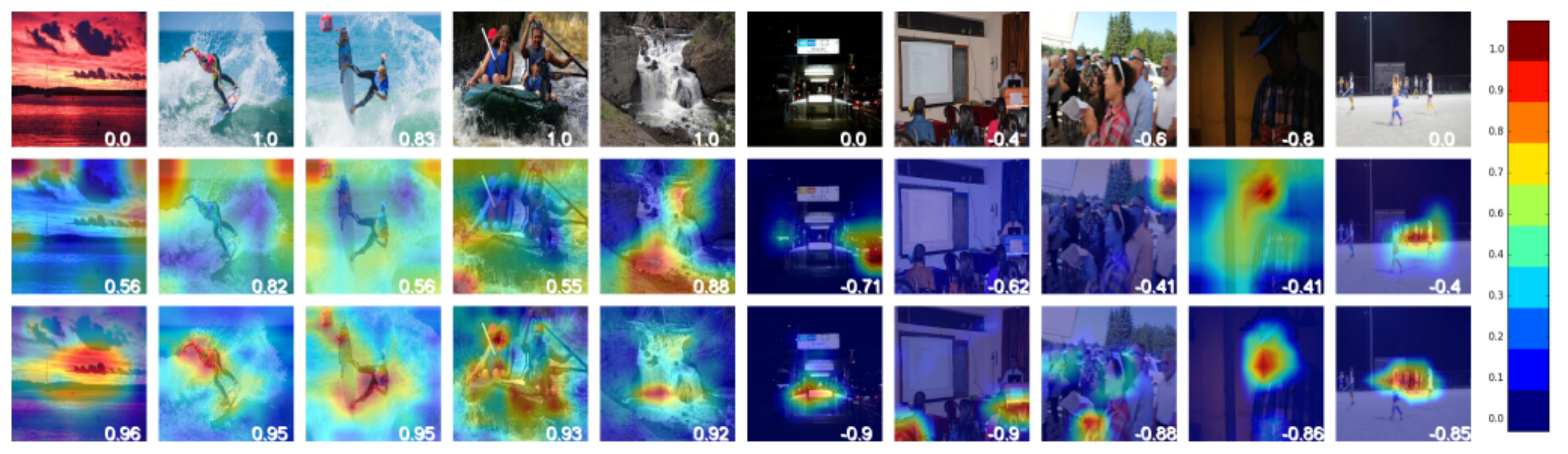}
  \caption{Content activation maps. First row: Original Images \small(marked with ground truth score at the bottom right\small), second row: Activation Maps from Kong \textit{et al.}~\cite{kong2016photo} model~\small(marked with predicted score from their given model\small), third row: Activation Maps from our method~\small(marked with our predicted score \small). Color-bar indicates the color encoding of activation.}
  \label{fig:content}
\end{figure*}
\section{Visualization}
As mentioned above we generate attribute activation maps for different attributes, to get their localized representations. Here we omit the following attributes, namely, emph{balancing element} and the \emph{rule of thirds}, as our model\textquotesingle s performance is very low for these attributes as shown in Table~\ref{tab:results}. For each attribute, we have analyzed the activation maps and present the insights in this section. For illustration purposes, We have selected ten samples for each attribute. Out of these ten samples, first five are the highest rated by our model, and the next five are the lowest rated. We have selected these samples from test samples \small($1000$\small) and not from the train samples. We also have included the activation maps from model given by Kong \textit{et al.}~\cite{kong2016photo} \small(Kong's model\small). These activation maps highlight the most important regions for the given attributes. We define these activation maps as ``gaze" of the model. 
\begin{figure*}
  \includegraphics[width=\textwidth]{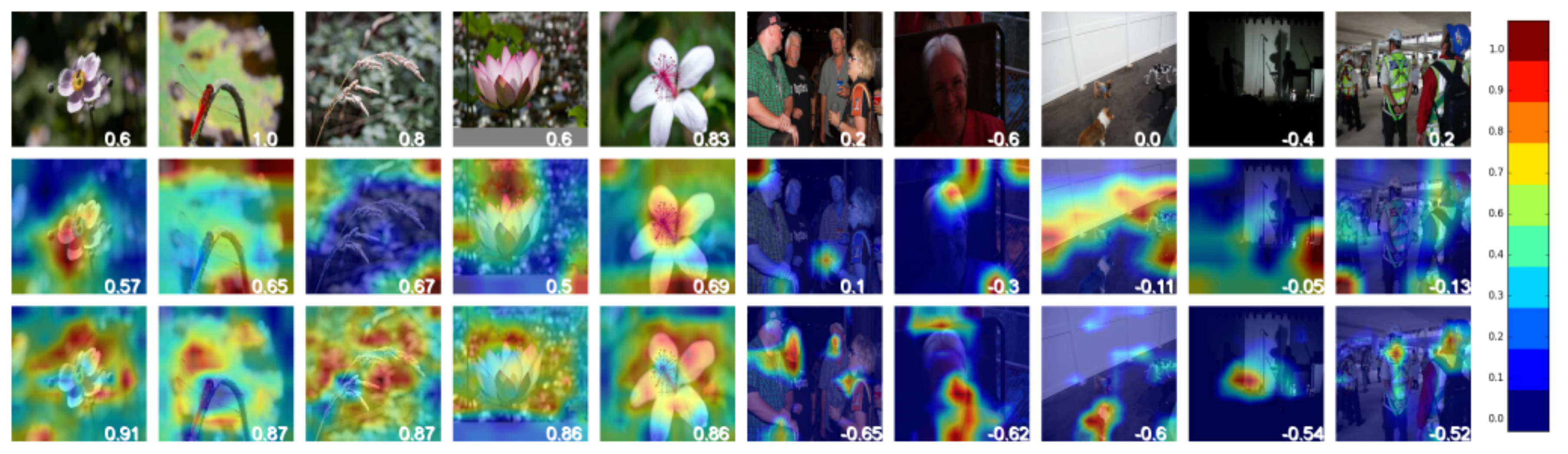}
  \caption{Depth of Field activation maps. First row: Original Images \small(marked with ground truth score at the bottom right\small), second row: Activation Maps from Kong \textit{et al.}~\cite{kong2016photo} model~\small(marked with predicted score from their given model\small), third row: Activation Maps from our method~\small(marked with our predicted score \small). Color-bar indicates the color encoding of activation.}
  \label{fig:DoF}
\end{figure*}
\begin{figure*}
  \includegraphics[width=\textwidth]{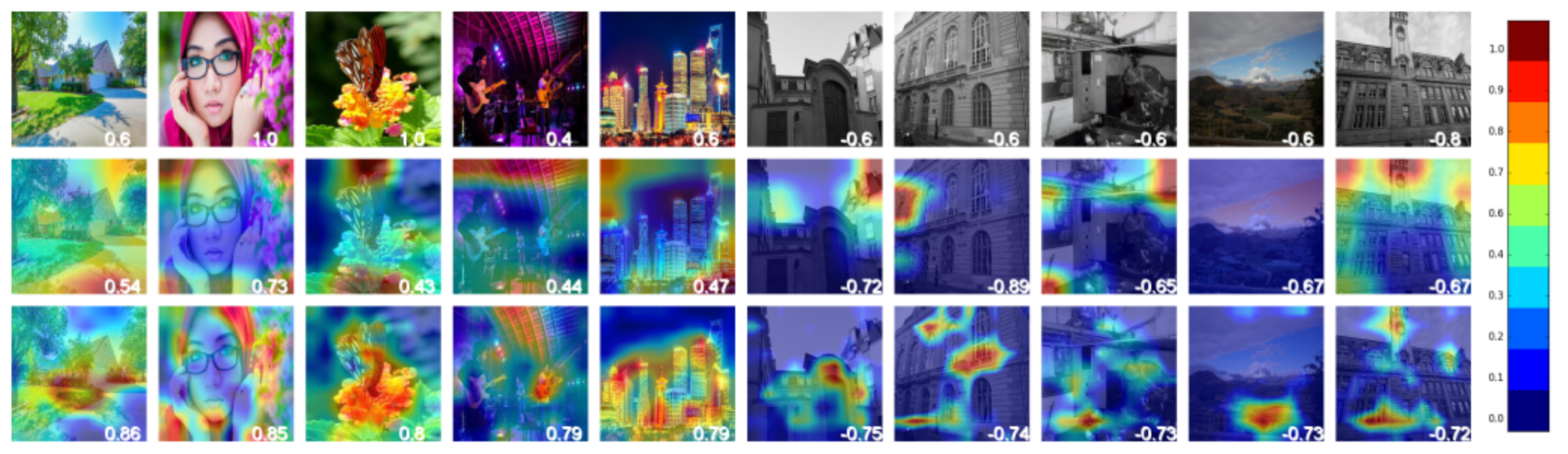}
  \caption{Vivid Color activation maps. First row: Original Images \small(marked with ground truth score at the bottom right\small), second row: Activation Maps from Kong \textit{et al.}~\cite{kong2016photo} model~\small(marked with predicted score from their given model\small), third row: Activation Maps from our method~\small(marked with our predicted score \small). Color-bar indicates the color encoding of activation.}
  \label{fig:VividColor}
\end{figure*}

\subsection{Object Emphasis}
By qualitative analysis of activation maps of object emphasis, it was observed that model gazes at the main object on the image. Even when the model predicts negative rating, i.e. object is not emphasized, the model searches for regions which contain objects Figure ~\ref{fig:object}. In comparison, activation maps from Kong's model are not always consistent as can be seen in the second row of activation maps in Figure~\ref{fig:object}. It showcases that our model has learned the object emphasis attribute as an attribute which is indeed related to objects. 
\subsection{Content}
Interestingness of content is significantly subjective and is a context-dependent attribute. However, if a model is trained on this attribute, one would expect the model would have maximum activation at the content of the image while making this judgment. If there exists a well-defined object in an image, then that object is considered as the content of the image, for e.g., $2^{nd}$ and $3^{rd}$ columns of Figure ~\ref{fig:content}. Further, it can be observed in these columns that our proposed approach is better at identifying the content than Kong's model. Without the presence of explicit objects, the content of the image is difficult to localize, for e.g. $1^{st}$ and $5^{th}$ columns of Figure ~\ref{fig:content}. As shown in Figure~\ref{fig:content}, our model's activation maps are maximally active at the content of the image. In comparison activation maps from Kong's models are not consistent.

\subsection{Depth of Field}
On analyzing the representations of shallow depth of field, it was observed that model looks for blurry regions near the main object of the image while making the judgment as showcased in Figure ~\ref{fig:DoF}. Shallow depth of field technique is used to make the subject of the photograph stand out from its background. The model's interpretation of it is in that direction. The images for which model has predicted the negative score on this attribute, the activation maps are random. Activation maps from Kong's model also showcase a similar behavior, these maps are more active at the corner of the images. 

\subsection{Vivid Color}
Vivid Color means the presence of bright and bold colors. The model's interpretation of this attribute seems to be along these lines. As shown in Figure ~\ref{fig:VividColor}, model gazes at vivid color areas while making the judgment about this attribute. For example, in $2^{nd}$ column of the Figure ~\ref{fig:VividColor} pink color of flowers and scarf, and in $3^{rd}$ column butterfly and flower were the most activated regions. 
Authors couldn't find any pattern in activation maps from Kong's model.  

\subsection{Light}
\begin{figure*}
  \includegraphics[width=\textwidth]{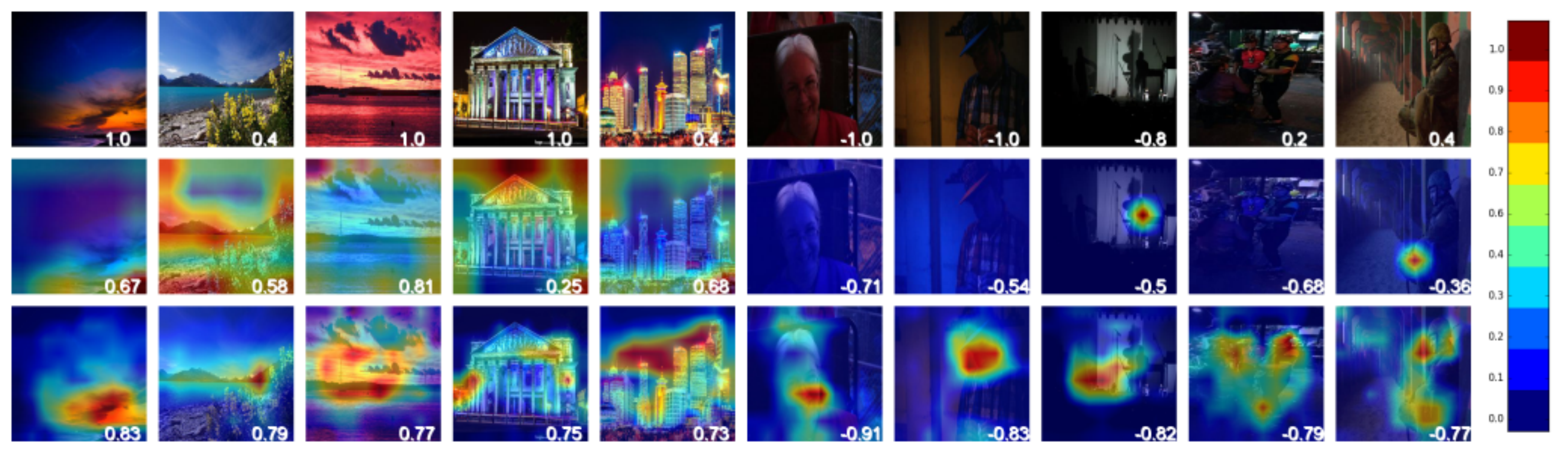}
  \caption{Light activation maps. First row: Original Images \small(marked with ground truth score at the bottom right\small), second row: Activation Maps from Kong \textit{et al.}~\cite{kong2016photo} model~\small(marked with predicted score from their given model\small), third row: Activation Maps from our method~\small(marked with our predicted score \small). Color-bar indicates the color encoding of activation.}
  \label{fig:Light}
\end{figure*}

Good Lighting is quite a challenging concept to grasp. It does not merely depend on the light in the photograph, but rather how that light complements the whole composition. As shown in Figure ~\ref{fig:Light}, most of the time model seems to look at bright light, or source of the light in the photograph. Although model\textquotesingle s behavior is consistent, its understanding of this attribute is incomplete. This was also evident in the low correlation ratings of our proposed model for this attribute, as reported in Table~\ref{tab:results}.  

\subsection{Color Harmony}
\begin{figure*}
  \includegraphics[width=\textwidth]{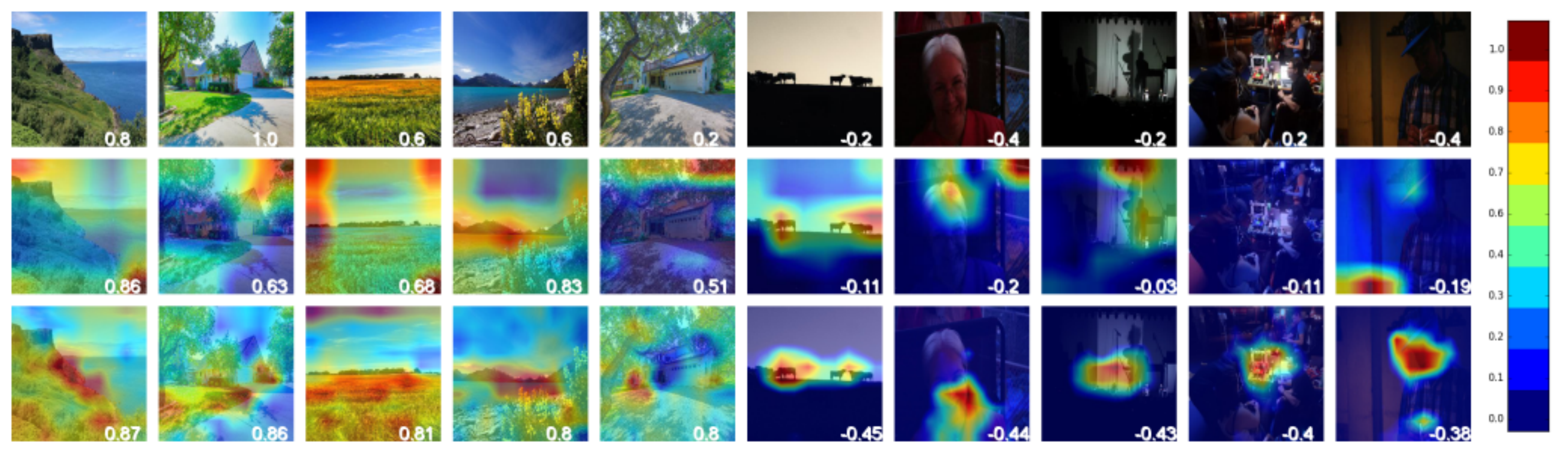}
  \caption{Color Harmony activation maps. First row: Original Images \small(marked with ground truth score at the bottom right\small), second row: Activation Maps from Kong \textit{et al.}~\cite{kong2016photo} model~\small(marked with predicted score from their given model\small), third row: Activation Maps from our method~\small(marked with our predicted score \small). Color-bar indicates the color encoding of activation.}
  \label{fig:ColorHarmony}
\end{figure*}

Although model\textquotesingle s performance is significant for this attribute, we could not find any consistent pattern in its activation maps. As color harmony is of many types, e.g., analogues, complementary, triadic; it is difficult to get a single representation pattern. For example, in the first example shown in Figure ~\ref{fig:ColorHarmony}, the green color of hills is in analogous harmony with blue color of water and sky; in the $3^{rd}$ example, brown sand color is in split complementary harmony with blue and green. The attribute activation maps for Color Harmony are shown in Figure ~\ref{fig:ColorHarmony}.  
\section{Conclusion}
In this paper, we have proposed deep convolution neural network (DCNN) architecture to learn aesthetic attributes. Results show that estimated scores of five aesthetic attributes (Interestingness of Content, Object emphasis, shallow Depth of Field, Vivid Color, and Color Harmony) correlate significantly with their respective ground truth scores. Whereas in the case of attributes such as Balancing Elements, Light and Rule of Thirds, the correlation is inferior. The activation maps corresponding to the learned aesthetic attributes such as object emphasis, content, depth of field and vivid color indicate that the model has acquired internal representation suitable to highlight these attributes automatically. However, for color harmony and light, the visualization maps were not consistent. 

Aesthetic judgment involves a degree of subjectivity. For example, in AADB the average correlation between the mean score and an individual's score for the overall aesthetic score is 0.67 ~\ref{tab:human_results}. Moreover, as reported by Kong \textit{et al}.~\cite{kong2016photo}, the model learned on a particular dataset might not work on a different dataset. Considering all these factors, empirical validity of aesthetic judgment models is still a challenge. We suggest that the visualization techniques presented in the current work is a step forward in that direction. Empirical validation could proceed by asking subjects to annotate the images (identifying the regions that correspond to different aesthetic attributes) and these empirical maps could in turn be compared with the predicted maps of the model. Such experiments need to be conducted in future to validate the current approach. 


\begin{acks}
The authors would like to thank Ms. Shruti Naik, and Mr. Yashaswi Verma of International Institute of Information Technology, Hyderabad, India for their  help in manuscript preparation.
\end{acks}

\bibliographystyle{ACM-Reference-Format}

\end{document}